\documentclass[letterpaper]{article} % DO NOT CHANGE THIS
\usepackage[submission]{aaai24}  % DO NOT CHANGE THIS
\usepackage{times}  % DO NOT CHANGE THIS
\usepackage{helvet}  % DO NOT CHANGE THIS
\usepackage{courier}  % DO NOT CHANGE THIS
\usepackage[hyphens]{url}  % DO NOT CHANGE THIS
\usepackage{graphicx} % DO NOT CHANGE THIS
\usepackage{natbib}  % DO NOT CHANGE THIS AND DO NOT ADD ANY OPTIONS TO IT
\usepackage{caption} % DO NOT CHANGE THIS AND DO NOT ADD ANY OPTIONS TO IT
\usepackage{algorithm}
\usepackage{algorithmic}
\usepackage{amsmath}
\usepackage{amssymb}
\usepackage{soul}
\usepackage{booktabs}
\usepackage{multirow}
\usepackage{colortbl}  %彩色表格需要加载的宏包
\definecolor{grey1}{RGB}{230, 230, 230}
\usepackage{xcolor}
\usepackage{subcaption}
\usepackage{tabulary}
\definecolor{Gray}{gray}{0.95}
\newcommand{\gr}[1]{{\textcolor{gray}{#1}}}
\newcommand{\gb}[1]{{\cellcolor{gray!18}{#1}}}
\newlength\savewidth
\definecolor{baselinecolor}{gray}{.95}

\usepackage{makecell}
\usepackage{newfloat}
\usepackage{listings}
\DeclareCaptionStyle{ruled}{labelfont=normalfont,labelsep=colon,strut=off} % DO NOT CHANGE THIS
\floatstyle{ruled}
\newfloat{listing}{tb}{lst}{}
\floatname{listing}{Listing}
\usepackage{hyperref}
\hypersetup{hidelinks,
        linkcolor=red,
	colorlinks=true,
	pdfstartview=Fit,
	breaklinks=true}

\title{Adaptive FSS: A Novel Few-Shot Segmentation Framework \\ via Prototype Enhancement}
\author{
    Jing Wang\textsuperscript{\rm 1,2}, 
    Jinagyun Li\textsuperscript{\rm 1,2}, Chen Chen\textsuperscript{\rm 3}, Yisi Zhang\textsuperscript{\rm 1,2}, \\Haoran Shen\textsuperscript{\rm 1,2}
    , Tianxiang Zhang\textsuperscript{\rm 1,2}\thanks{Corresponding Author.}
}
\affiliations{
    \textsuperscript{\rm 1}School of Automation and Electrical Engineering, University of Science and Technology Beijing, Beijing, China\\
    \textsuperscript{\rm 2} Key Laboratory of Knowledge Automation for Industrial Processes, Ministry of Education, Beijing, China\\
    \textsuperscript{\rm 3} Center for Research in Computer Vision, University of Central Florida, Orlando, USA\\
    \url{https://jingw193.github.io/AdaptiveFSS/} \\
    m202120718@xs.ustb.edu.cn, leejy@ustb.edu.cn, chen.chen@crcv.ucf.edu, \\ 
    \{m202210579, m202220738\}@xs.ustb.edu.cn, txzhang@ustb.edu.cn.

}

\usepackage{bibentry}
\begin{document}

\maketitle

\begin{abstract}
The Few-Shot Segmentation (FSS) aims to accomplish the novel class segmentation task with a few annotated images. Current FSS research based on meta-learning focuses on designing a complex interaction mechanism between the query and support feature. However, unlike humans who can rapidly learn new things from limited samples, the existing approach relies solely on fixed feature matching to tackle new tasks, lacking adaptability. In this paper, we propose a novel framework based on the adapter mechanism, namely Adaptive FSS, which can efficiently adapt the existing FSS model to the novel classes. In detail, we design the Prototype Adaptive Module (PAM), which utilizes accurate category information provided by the support set to derive class prototypes, enhancing class-specific information in the multi-stage representation. In addition, our approach is compatible with diverse FSS methods with different backbones by simply inserting PAM between the layers of the encoder. Experiments demonstrate that our method effectively improves the performance of the FSS models (e.g., MSANet, HDMNet, FPTrans, and DCAMA) and achieves new state-of-the-art (SOTA) results (i.e., 72.4\% and 79.1\% mIoU on PASCAL-5$^i$ 1-shot and 5-shot settings, 52.7\% and 60.0\% mIoU on COCO-20$^i$ 1-shot and 5-shot settings).

\end{abstract}

\section{Introduction}
As one of the fundamental tasks in the field of computer vision, semantic segmentation has achieved significant improvement driven by the rapid increase of data scale.
However, acquiring detailed, pixel-level annotations for images is a well-known challenge, both time-consuming and costly \cite{everingham2010pascal, lin2014microsoft}. This complexity adds to the difficulty for models in learning about new categories. To address this, few-shot segmentation (FSS) is proposed to learn new concepts on a few labeled samples (i.e. support images), realizing new class segmentation on unlabeled images (i.e. query images).

% two-branch

\begin{figure}[!t]
    \centering
    \includegraphics[width=0.48\textwidth]{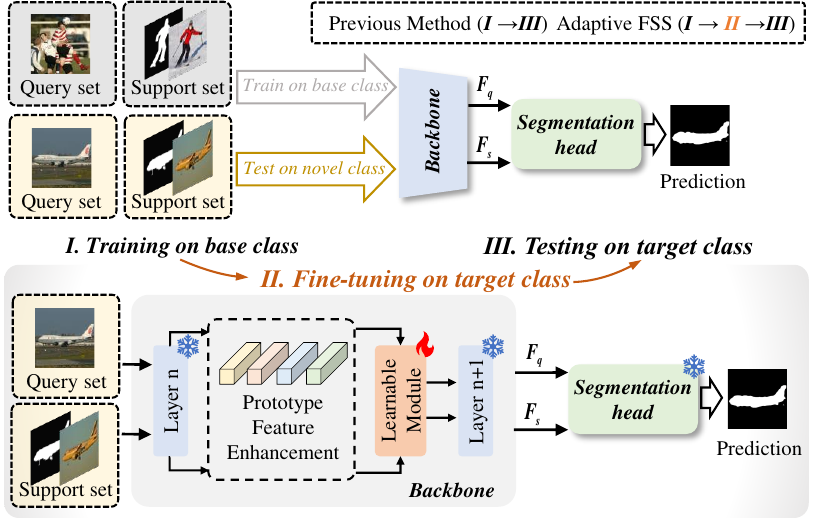}
    % \vspace{-35pt}
    \caption{The overview of our Adaptive FSS. Previous works generally train the FSS model on the base classes and directly evaluate it on novel classes. In our framework, we insert the \textbf{P}rototype \textbf{A}daptive \textbf{M}odule (PAM), conducting a fine-tuning step before testing to effectively adapt the model to novel classes through prototype enhancement.}

    \vspace{-10pt}
    \label{fig1}
\end{figure}

Recently, most FSS researches \cite{kang2022integrative,yang2020prototype,lang2022learning} employ a meta-learning episodic training strategy and typically focus on elaborately designing a base segmentation model, which includes sophisticated feature interaction mechanisms \cite{zhang2021few,iqbal2022msanet} between query images and support images. Specifically, in each training episode, a query image and a support set are sampled to imitate the situation of segmenting a novel category. During meta-testing, the base segmentation model predicts the query mask without changing any parameters as illustrated in the upper portion of Fig. \ref{fig1}. A fundamental challenge of this pipeline is how to rapidly learn task-specific information corresponding to new concepts and adapt the model to novel classes using limited labeled samples. Simultaneously, in the few-shot classification task \cite{dhillon2019baseline,kang2021relational}, a common approach to achieving the above objectives and significant performance improvements is to pre-train the entire model and then fine-tune the specific head for new tasks.

\ul{However, designing an effective fine-tuning framework suitable for existing FSS methods presents the following challenges given limited samples.}
Firstly, determining the appropriate number and optimal placement of updated parameters in the FSS model is difficult for ensuring the effectiveness of fine-tuning. For example, in the meta-learning pipeline of FSS, the primary objective of the segmentation head is mainly to distinguish the foreground and the background without a strong relation to the specificity of the category. For dense prediction FSS tasks, the adaptability of fine-tuning a task-specific head like in classification based on image-level semantic matching may be insufficient (related experimental results can be seen in Table \ref{SOTA_result}). 
On the other hand, fine-tuning the entire model (i.e., encoder or decoder) is prone to over-fitting in the case of limited data.

Secondly, diverse backbones (i.e., ResNet \cite{he2016deep}, ViT  \cite{dosovitskiy2020image}, Swin Transformer \cite{liu2021swin}) and various types of decoders are employed in existing FSS methods \cite{peng2023hierarchical,zhang2022feature,shi2022dense}, which is an obstacle to design a general fine-tuning architecture.
Besides, fine-tuning parameters in the original model will lead to catastrophic forgetting and compression of knowledge about base classes.

Currently, adapter \cite{houlsby2019parameter, chen2022adaptformer} is a commonly used parameter-efficient transfer learning method, and is utilized to accommodate few-shot classifiers \cite{zhang2022tip,li2022cross} to new tasks and domains.
During training, it only updates the parameters in the additionally inserted adapter module and freezes the rest of the model. By flexibly adjusting the parameter scale, insertion position, and design ideas of the adapter, the model can effectively adapt to new tasks while mitigating over-fitting and preventing catastrophic forgetting. Therefore, adapter tuning may be an appropriate way to solve the aforementioned problems.
\ul{However, for the FSS task, the existing adapter cannot effectively exploit the provided support set to extract class-specific features. Besides, it is difficult to model general representations for different classes with limited samples.} 

In this paper, we propose the Adaptive FSS framework, exploring the potential of the fine-tuning process based on the adapter mechanism in the FSS task. 
\emph{To address the challenge of applying the existing adapter module to the FSS task, we meticulously design a powerful module named the Prototype Adaptive Module (PAM), as shown in Fig. \ref{fig1}.}
Specifically, it consists of a prototype enhancement module (PEM) and a learnable adaptive module (LAM). 
In the PEM, the support set is utilized to encode a class prototype, enhancing the information associated with the new class in the features. 
Besides, the class prototype bank updated by momentum is defined to improve the quality of prototypes and acquire the general class representation.
For LAM design, we employ a simple set of projection layers to further model the task-specific information. 
From the perspective of feature adaptation, our approach only inserts the PAM into the backbone to obtain multi-stage category-specific features for improving model adaptability.
Importantly, this method does not necessitate specialized decoder designs, making it universally applicable to any existing FSS model with various backbones. By fine-tuning PAM accounting for a small proportion (0.5\% on average) of the whole network, the base segmentation model can rapidly adapt to the new category.

We conduct extensive experiments on two benchmarks (i.e. PASCAL-5$^i$ \cite{everingham2010pascal} and COCO-20$^i$ \cite{lin2014microsoft}) based on four well-performed FSS models (i.e. MSANet \cite{iqbal2022msanet}, HDMNet \cite{peng2023hierarchical}, FPTrans \cite{zhang2022feature}, and DCAMA \cite{shi2022dense}) to prove the effectiveness of the proposed Adaptive FSS. The results demonstrate that our approach is a feasible and effective solution for enhancing new class adaption to boost the SOTA performance in FSS tasks. Our main contributions could be summarized as follows:

\begin{itemize}
    \item We propose a novel framework Adaptive FSS on applying the adapter mechanism to the few-shot segmentation task. This general architecture can be integrated into various FSS methods, facilitating effective adaptation of the base segmentation model to novel classes.
    \item We design a novel Prototype Adaptive Module (PAM) to realize the enhancement of refined features for specific categories during fine-tuning. Moreover, it is plug-and-play and well-suitable for the FSS task.
    \item The experimental results on two benchmark datasets demonstrate that our proposed method achieves superior performance over SOTA approaches (with an average $\uparrow$ 2.8\% mIoU on PASCAL-5$^i$ in the 1-shot setting, requiring only a 0.5\% parameters increase). 
\end{itemize}

\section{Related Work}

\subsection{Few-Shot Learning}
Few-Shot Learning (FSL) aims to enable models to learn and adapt, generalizing to novel domains based on the hints of a few labeled samples. Existing FSL researches mainly focus on image classification \cite{hou2019cross, kang2021relational, zheng2022few} and other visual tasks \cite{kang2023distilling}, and have led to several primary categories of solutions in this field, including fine-tuning \cite{ravi2016optimization, Finn2017ModelAgnosticMF}, metric learning \cite{Sung2017LearningTC, Yoon2019TapNetNN, Schwartz2018RepMetRM} and meta-learning \cite{Zhang2022HGMetaGM}. Among them, fine-tuning-based methods involve training models on a substantial amount of source samples and fine-tuning them on a small set of task-specific samples.

\subsection{Few-Shot Segmentation}
The mainstream training pipeline for few-shot segmentation(FSS) is based on episodic training, which is a typical form of meta-learning. Early methods \cite{dong2018few, tian2020differentiable,rakelly2018conditional, zhang2020sg, zhang2021self} for FSS generally follow a classic dual-branch structure proposed by \cite{shaban2017one}, the pioneering work of FSS. The trend of recent methods is the single-branch structure based on prototypical network \cite{wang2019panet, xie2021scale, tian2020prior, yang2020prototype, kang2022integrative}. 

Rather than constructing a prototype feature extractor, some works \cite{vinyals2016matching, nguyen2019feature, zhang2019canet, yang2020brinet, lu2021simpler, liu2021few} are devoted to capturing correspondence between the query image and support set. HSNet \cite{min2021hypercorrelation} exploits multi-level feature relevance and efficient 4D convolution to filter the query-support correlation map. APANet \cite{chen2021apanet} introduces an adaptive prototype representation to regulate incomplete feature interaction. 
Yet, designs based on meta-learning suffer from severe loss of spatial structure and limitations of the inherent priors. CRNet \cite{liu2020crnet} opens up another way, proposing a combination of conditional network, Siamese network encoder, cross-inference module, and mask refinement to achieve FSS. RePRI \cite{boudiaf2021few} abandons the strong assumption in meta-learning that base set and novel set have similar distributions and resorts to simply supervised base training by cross-entropy loss.

However, current few-shot segmentation approaches are limited by modeling the novel class-specific information and focus on the research of fixed feature matching mechanism between query and support. To this end, we propose a framework to implement FSS through Parameter-Efficient-Tuning, empowering models to learn new classes in a stable and parameter-efficient way.

\begin{figure*}[!tp]
    \centering
    \includegraphics[width=1\textwidth]{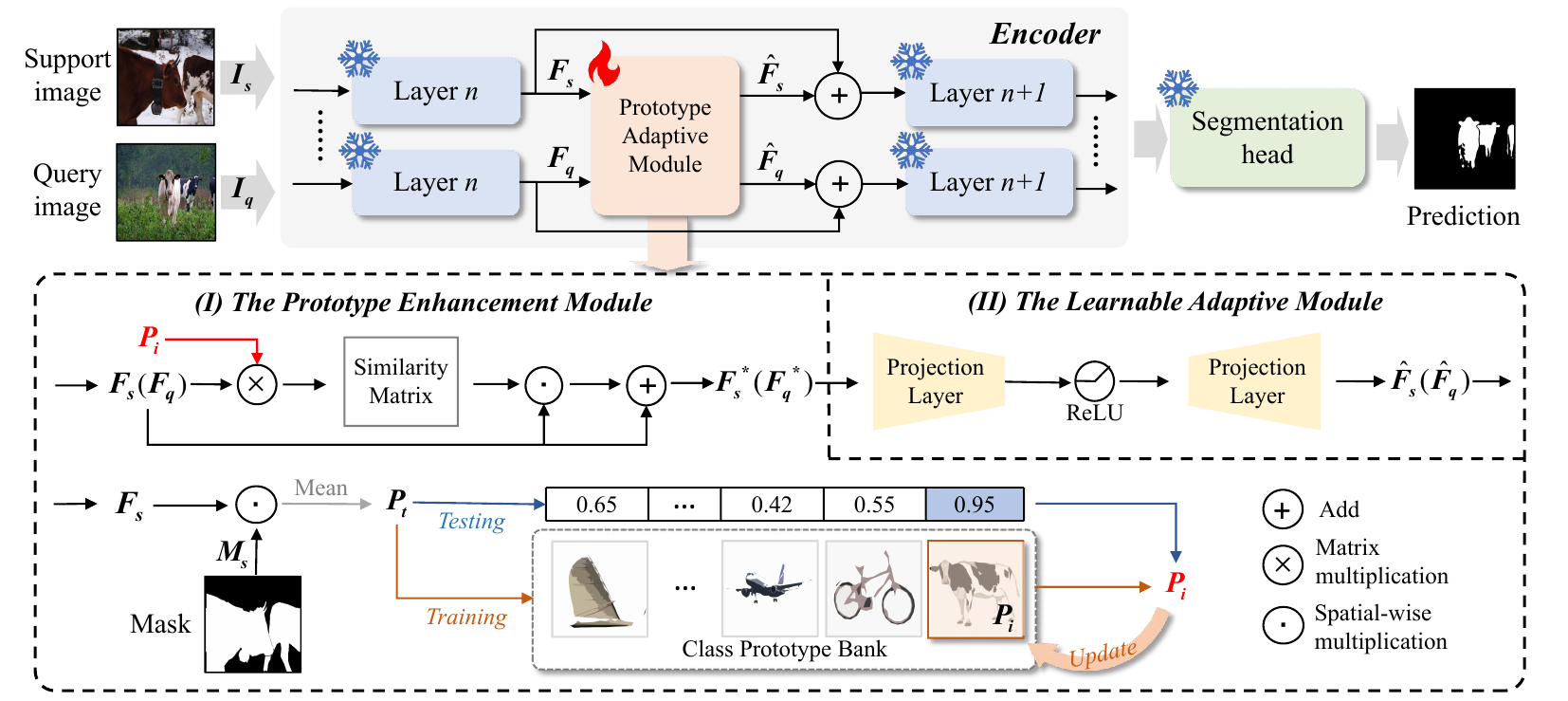}
    \vspace{-10pt}
    \caption{The overall architecture of our proposed Adaptive FSS. Given a support set $\{I_s, M_s\}$, the image $I_s$ is fed into the encoder and generates feature $F_s$ ($F_q$). In each PAM, with calculation between $F_s$ and mask $M_s$, the temporary prototype $P_t$ is first obtained to select prototype $P_i$ and update the bank. After that, the corresponding class prototype $P_i$ and feature $F_s$ ($F_q$) are combined to generate the class-specific feature $F_s^*$ ($F_q^*$). Finally, $F_s^*$ ($F_q^*$) is sent into the Learnable Adaptive Module, leading to the acquired $\hat{F_s}$ ($\hat{F_q}$), which are injected into the encoder.}
    \label{fig_method}
    \vspace{-10pt}
\end{figure*}

\subsection{Parameter-Efficient Transfer Learning}
In the FSS task, learning new classes using standard fine-tuning methods is prone to over-fitting or catastrophic forgetting. To address this, the use of Parameter-Efficient Tuning (PET) technique emerges as an effective solution. There has been a growing demand for PET and several alternative directions have emerged, including Adapter \cite{houlsby2019parameter, hu2023llm}, Prompt-tuning \cite{zhu2022continual, liu2022p}, Prefix-tuning \cite{li2021prefix}, and LoRA \cite{hu2021lora}. In computer vision, Adapter \cite{chen2022adaptformer, chen2022conv, pan2022st} and Prompt-tuning \cite{wang2021list, jia2022visual} exhibit great performance in transfer learning.

In recent years, adapter method has also been introduced into the FSL task. Researchers have explored adapting models to unseen classes by attaching sets of additive parameters called adapters into original models and fine-tuning these sets. FSL algorithms based on adapter-tuning \cite{li2022cross, gao2021clip, zhang2022tip} represent striking classification accuracy and universal.

\section{Methodology}
\label{Methodology}
This section first introduces the primary setup of regular FSS (Few-shot segmentation). Then, the details of the proposed Adaptive FSS architecture are presented. After that, the Prototype Adaptive Module (PAM) structure is elaborated from the two components of the Prototype Enhancement Module (PEM) and the Learnable Adaptive Module (LAM).

\subsection{Problem Setup}
\textbf{Few-shot segmentation Setup.} 
 Few-shot segmentation (FSS) focuses on tackling the semantic segmentation problem on novel classes with only the corresponding few samples. For an n-way k-shot FSS task, current research \cite{zhang2021few} widely adopts the meta-learning paradigm called episodic training \cite{vinyals2016matching}, where each episode is associated with a single category becoming a 1-way k-shot task. In general, the dataset is first split into $D_{tr}$ and $D_{ts}$ for training and testing. Further, $D_{tr}$ and $D_{ts}$ are divided into $\{D_{tr}^{0}, D_{tr}^{1}, ..., D_{tr}^{c-1}\}$ and $\{D_{ts}^{0}, D_{ts}^{1}, ..., D_{ts}^{c-1}\}$ ($c$ denotes the number of classes) by classes respectively. For training each episode, a query sample $\{I_q, M_q\}$ and $K$ support samples $\{I_s^k, M_s^k\}_{k=1}^K$ are selected from $\{D_{tr}^{i}\}$ and the model is expected to predict query mask $M_q$ according to $\{I_s^k, M_s^k\}_{k=1}^K$ and $I_q$. For testing, the query-support images are sampled from $D_{ts}^{j}$ ($j \neq i$) to achieve inference of novel classes. In this paper, we follow this popular episodic training and testing scheme.

\subsection{Adaptive FSS Framework}
\subsubsection{Overview.} 
% In this paper, we explore a generic FSS framework based on novel class fine-tuning. 
An overview of our proposed Adaptive FSS framework is presented in Fig. \ref{fig_method}. Given a query image $I_q$ and a support set $\{I_s^k, M_s^k\}_{k=1}^K$, the encoder first extracts query feature $F_q$ and support feature $F_s$ as the previous methods. 
Then, $F_s$, $M_s$, and $F_q$ are input to our proposed PAM to obtain the class-specific features $F_s^*$ and $F_q^*$ by PEM. Further, we feed the $F_s^*$ and $F_q^*$ into LAM to learn special information for novel tasks, generating $\hat{F_s}$ and $\hat{F_q}$. 
After that, $\hat{F_s}$ and $\hat{F_q}$ are injected into the original feature $F_s$ and $F_q$, which is employed in the downstream decoder to achieve more precise segmentation. 
When fine-tuning, we only update the parameters of the PAM and freeze the rest of the network, making the base segmentation model adapt to the new category efficiently.

\subsubsection{Prototype Enhancement Module.}
\label{Prototype_Enhancement_Module}
It consists of prototype generation to get high-quality class prototypes and feature enhancement to reinforce class-specific information according to prototypes, which are elaborated as follows.

\noindent \emph{\textbf{Prototype Generation}.} As shown in Fig. \ref{fig_method}, for a n-way k-shot task, the support feature $F_s \in \mathbb{R}^{k \times d \times h \times w }$ and mask $M_s \in \mathbb{R}^{k \times h \times w}$ ($k$ denotes the number of support samples, $d$ presents the corresponding feature dimension, $h$ and $w$ denote feature height and width. $M_s$ is down-sampled to the same resolution as the features.) are utilized to obtain higher-quality category prototypes.
Specifically, we first define a class prototype bank $P = \{P_1, P_2, ... P_n\} \in \mathbb{R}^{n \times d}$ in each PAM, where $n$ denotes the number of novel classes and $P_{i \in [1:n]} \in \mathbb{R}^{d}$ presents the prototype of target class. 
Then, a temporary prototype $P_t \in \mathbb{R}^{d}$ of the target class is obtained by calculating the spatial-wise multiplication between the support feature $F_s$ and masks $M_s$ and averaging nonzero feature embedding. The above mathematical expression is defined as follows:
\begin{equation}
    \label{temp_prototype}
    P_t = \mathrm{Mean}(F_s \circ M_s), 
\end{equation}
where $\circ$ presents spatial-wise multiplication and $\mathrm{Mean}$ denotes calculating the mean in the spatial dimension only on the nonzero position of $M_s$. This way can avoid the influence of the proportion difference of the object on the whole image.
With the temporary prototype $P_t$, we update and select the corresponding class prototype $P_i$ from the previous class prototype bank $P$ during training and testing respectively. 
In the training phase, when adapting target class $i$, $P_i$ is located from $P$ according to $i$ for precisely improving the representational quality of the class prototype. Then, we momentum update $P_i$ by:
\begin{equation}
    \label{momentum_update}
    P_i = (1 - \alpha) \times l_2(P_t) + \alpha \times l_2(P_i), 
\end{equation}
where $l_2$ and $\alpha$ denote \emph{L$_2$} normalization and momentum ratio respectively. In the testing phase, the prototype $P_i$ is selected by similarity matching between $P_t$ and $P$. 
\begin{equation}
    \label{testing_select}
    P_i = P_{\mathrm{argmax}(l_2(P) \cdot l_2(P_t))}, 
\end{equation}
In this manner, each $P_i$ can well represent the semantics of the corresponding category in the current feature representation space.

\noindent \emph{\textbf{Feature Enhancement}.}
As shown in Fig. \ref{fig_method}, given a $P_i$, we can accurately enhance the part of the feature that associates with the new class, making model easier to distinguish between the foreground and background.
Firstly, we calculate the similarity map $S_s \in \mathbb{R}^{k \times h\times w}$ between $F_s$ and $P_i$, namely 
\begin{equation}
    \centering
    \label{enhancement_similarity}
    \begin{aligned} 
    S_s & = l_2(F_s) \cdot l_2(P_i), \\ 
    \mathrm{ReLU6}(x) & = \mathrm{Min}(6, \mathrm{Max}(0, x)), \\
    E_m & = \mathrm{ReLU6}( S_s \times \sqrt{d}), \\
    \end{aligned}
\end{equation}
where \emph{L$_2$} normalization and dot multiply are employed at $d$ dimension. By utilizing the ReLU6, dissimilar positions are suppressed and similar points are retained while avoiding the impact of excessive values. With the enhancement matrix $E_m$, we can generate the class-specific feature $F_s^*$:
\begin{equation}
    \label{enhancement_feature}
    F_s^* = E_m \circ F_s + F_s. 
\end{equation}
To be emphasized, the query $F_q$ feature is also enhanced by the above process with the same prototype $P_i$. The above only takes the support feature $F_s$ as an example to illustrate.

\subsubsection{The Learnable Adaptive Module.}
\label{Learnable_Adaptive_Module}
After the above process, the enhanced feature $F_s^*$ and $F_q^*$ can be obtained. However, the ability of the model to encode task-specific information is insufficient. Moreover, there is a problem of distribution differences between enhanced features $F_s^*$ ($F_q^*$) and original features $F_s$ ($F_q$) and it will increase cumulatively with the number of layers deepens.
% 主动建模能力不足
Therefore, a learnable module is adopted to solve these problems in our PAM. Since the available samples for training are very limited, the parameters of this module are very few that can be ignored compared with the entire model to alleviate the over-fitting phenomenon.
Specifically, it includes a down-projection linear layer with parameters $W_{down} \in \mathbb{R}^{d \times \tfrac{d}{\gamma}}$ for compressing feature dimensions and an up-projection linear layer with parameters $W_{up} \in \mathbb{R}^{\tfrac{d}{\gamma} \times d}$ to recover feature dimensions. $\gamma$ presents the hidden dimensions ratio and a ReLU layer is placed between two layers to complement the non-linear properties. The task-specific features $\hat{F_s}$ and $\hat{F_q}$ can be obtained as
\begin{equation}
    \label{adapter}
    \begin{aligned} 
    \hat{F_s} = \mathrm{ReLU}(F_s^* \cdot  W_{down}) \cdot  W_{up}, \\ 
    \hat{F_q} = \mathrm{ReLU}(F_q^* \cdot  W_{down}) \cdot  W_{up}. 
    \end{aligned}
\end{equation}
Then, we inject features $\hat{F_s}$ and $\hat{F_q}$ into the encoder as shown in Fig. \ref{fig_method}. This process of injection is written as:
\begin{equation}
    \label{inject}
    \begin{aligned} 
    F_s = \hat{F_s} \times \beta + F_s, \\ 
    F_q = \hat{F_q} \times \beta + F_q, \\
    \end{aligned}
\end{equation}
where $\beta$ denotes the scaling factor.

\begin{table*}[!t]
  \centering
  \footnotesize
  \setlength\tabcolsep{1.9pt}%
  \begin{tabular}{c|l|c|cccc|c|cccc|c}
    \toprule[1.0pt]
    \multicolumn{13}{c}{\textbf{(a) PASCAL-5$^i$}}  \\
    % \hline
    \multirow{2}{*}{Backbone} & \multirow{2}{*}{Method} & \multirow{2}{*}{Params} & \multicolumn{5}{c|}{1-Shot} & \multicolumn{5}{c}{5-Shot} \\
    \cmidrule{4-13}
      &  & & Fold-0 & Fold-1 & Fold-2 & Fold-3 & mIoU\% & Fold-0 & Fold-1 & Fold-2 & Fold-3 & mIoU\% \\
    \midrule
     \multirow{6}{*}{ResNet 50} 
    
     &  \gr{BAM~\cite{lang2022learning}} &  \gr{51.63M} &
     \gr{69.0} &  \gr{73.6} & \gr{67.6} &  \gr{61.1}&
     \gr{67.8} & \gr{70.6} &  \gr{75.1} &  \gr{70.8} &  \gr{67.2} &  \gr{70.9} \\

     &  \gr{MIANet~\cite{yang2023mianet}} & \gr{59.01M} &
     \gr{68.5} &  \gr{75.8} & \gr{67.5} &  \gr{63.2}&
     \gr{68.7} & \gr{70.2} &  \gr{77.4} &  \gr{70.0} &  \gr{68.8} &  \gr{71.6} \\

     &  \gr{HDMNet~\cite{peng2023hierarchical}} & \gr{51.40M} &
     \gr{71.0} &  \gr{75.4} & \gr{68.9} &  \gr{62.1}&
     \gr{69.4} & \gr{71.3} &  \gr{76.2} &  \gr{71.3} &  \gr{68.5} &  \gr{71.8} \\

     &  \gr{MSANet~\shortcite{iqbal2022msanet}} & \gr{52.37M} & \gr{69.3} &  \gr{74.6} & \gr{67.8} &  \gr{62.4} & \gr{68.5} & \gr{72.7} &  \gr{76.3} &  \gr{73.5} &  \gr{67.9} &  \gr{72.6} \\
     &  MSANet* & 52.37M & 69.9 &  74.9 & 64.8 &  61.3 & 67.7 & 74.0 &  76.4 &  69.8 &  68.0 &  72.1 \\
     & \gb{MSANet + \textcolor{blue}{\textbf{Ours}}} & \gb{0.53M} & \gb{\textbf{71.1}} & \gb{\textbf{75.5}} & \gb{\textbf{67.0}} & \gb{\textbf{64.5}} & \gb{\textbf{69.5(+1.8)}} & \gb{\textbf{74.7}} & \gb{\textbf{78.0}} & \gb{\textbf{75.3}} & \gb{\textbf{70.8}} & \gb{\textbf{74.7(+2.6)}} \\
    \midrule
    \multirow{5}{*}{DeiT-B/16} 

    &  \gr{FPTrans~\cite{zhang2022feature}} & \gr{174.17M} & \gr{72.3} &  \gr{70.6} &  \gr{68.3} &  \gr{64.1} &  \gr{68.8} &  \gr{76.7} &  \gr{79.0} &  \gr{81.0} &  \gr{75.1} &  \gr{78.0} \\
    &  FPTrans* & 174.17M & 73.4 &  70.3 &  67.8 &  63.7 &  68.8 &  76.6 &  79.3 &  79.4 &  74.5 &  77.5\\

    & FPTrans + Adaptformer \gr{\shortcite{chen2022adaptformer}} &0.49M & 74.0 & 70.6 & 67.7 & 66.2 & 69.8(+1.0) & 77.1 & 79.8 & 80.0 & 74.7 & 77.9(+0.4)\\
    & FPTrans + VPT \gr{\cite{jia2022visual}} &0.25M & 73.8 & 70.6 & 67.7 & 64.2 &  69.1(+0.3) & 76.8 & 79.6 & 80.0 & 75.1 &  77.9(+0.4)\\
     & \gb{FPTrans + \textcolor{blue}{\textbf{Ours}}} & \gb{0.45M} &\gb{\textbf{74.1}} & \gb{\textbf{73.9}} & \gb{\textbf{71.3}} & \gb{\textbf{69.8}} & \gb{\textbf{72.3(+3.5)}} & \gb{\textbf{77.6}} & \gb{\textbf{80.6}} & \gb{\textbf{81.5}} & \gb{\textbf{76.5}} & \gb{\textbf{79.1(+1.6)}} \\
    \midrule
    \multirow{4}{*}{Swin-B} 
    
    & DCAMA\gr{~\cite{shi2022dense}} & 92.97M &  72.2  & 73.8 &  64.3 &  67.1 & 69.3 & 75.7 &  77.1 &  72.0 &  74.8 &  74.9 \\
   & DCAMA + F-Decoder & 5.07M & 73.2 & 73.7 & 65.8 & 67.5 & 70.1(+0.8) & 76.3 & 77.6 & 71.8 & 74.7 & 75.1(+0.2) \\
   & DCAMA + F-Head & 290 & 72.5 & 73.7 & 64.4 & 67.0 & 69.4(+0.1) & 75.8 & 77.2 & 72.1 & 74.8 & 75.0(+0.1) \\
     & \gb{DCAMA + \textcolor{blue}{\textbf{Ours}}} & \gb{0.23M} & \gb{\textbf{74.3}} & \gb{\textbf{74.9}} & \gb{\textbf{70.5}} & \gb{\textbf{69.8}} & \gb{\textbf{72.4(+3.1)}} & \gb{\textbf{76.8}} & \gb{\textbf{78.8}} & \gb{\textbf{74.3}} & \gb{\textbf{77.1}} & \gb{\textbf{76.7(+1.8)}} \\
  \end{tabular}
  \setlength\tabcolsep{1.9pt}%
      \begin{tabular}{c|l|c|cccc|c|cccc|c}
    \toprule[1.0pt]
    \multicolumn{13}{c}{\textbf{(b) COCO-20$^i$}}  \\
    \multirow{2}{*}{Backbone} & \multirow{2}{*}{Method} & \multirow{2}{*}{Params} & \multicolumn{5}{c|}{1-Shot} & \multicolumn{5}{c}{5-Shot} \\
    \cmidrule{4-13}
      & & & Fold-0 & Fold-1 & Fold-2 & Fold-3 &  mIoU\% & Fold-0 & Fold-1 & Fold-2 & Fold-3 &  mIoU\% \\
    \midrule
    \multirow{9}{*}{ResNet 50}
    
     &  \gr{BAM~\cite{lang2022learning}} & \gr{51.63M} & 
     \gr{43.4} &  \gr{50.6} & \gr{47.5} &  \gr{43.4}&
     \gr{46.2} & \gr{49.3} &  \gr{54.2} &  \gr{51.6} &  \gr{49.6} &  \gr{51.2} \\

     &  \gr{MIANet~\cite{yang2023mianet}} & \gr{59.01M} & 
     \gr{42.5} &  \gr{53.0} & \gr{47.8} &  \gr{47.4}&
     \gr{47.7} & \gr{45.8} &  \gr{58.2} &  \gr{51.3} &  \gr{51.9} &  \gr{51.7} \\

    &  \gr{MSANet \shortcite{iqbal2022msanet}} & \gr{52.37M} & \gr{45.7} &  \gr{54.1} & \gr{45.9} &  \gr{46.4}&
     \gr{48.0} & \gr{50.3} &  \gr{60.9} &  \gr{53.0} &  \gr{50.5} &  \gr{53.7} \\
    &  MSANet* & 52.37M &  41.9 &  53.1 & 45.5 &  46.7 &  46.8 &  46.7 &  60.3 &  53.1 &  50.4 &  52.6 \\
     & \gb{MSANet + \textcolor{blue}{\textbf{Ours}}} & \gb{0.53M} & \gb{\textbf{44.1}} & \gb{\textbf{55.0}} & \gb{\textbf{46.5}} & \gb{\textbf{48.5}} & \gb{\textbf{48.5(+1.7)}} & \gb{\textbf{48.1}} & \gb{\textbf{60.8}} & \gb{\textbf{54.8}} & \gb{\textbf{51.9}} & \gb{\textbf{53.9(+1.3)}} \\
    \cmidrule{2-13}
    &  \gr{HDMNet~\cite{peng2023hierarchical}} & \gr{51.40M} & 
     \gr{43.8} &  \gr{55.3} & \gr{51.6} &  \gr{49.4}&
     \gr{50.0} & \gr{50.6} &  \gr{61.6} &  \gr{55.7} &  \gr{56.0} &  \gr{56.0} \\
     &  HDMNet* & 51.40M & 44.0  & 55.1  & 50.1 &  48.7 &  49.5 & 52.5  & 64.5 &  55.2 &  55.1 &  56.8 \\
    & \gb{HDMNet + \textcolor{blue}{\textbf{Ours}}}  & \gb{0.53M} & \gb{\textbf{44.9}} & \gb{\textbf{56.7}} & \gb{\textbf{51.4}} & \gb{\textbf{49.6}} & \gb{\textbf{50.7(+1.2)}} & \gb{\textbf{53.0}} & \gb{\textbf{66.4}} & \gb{\textbf{56.1}} & \gb{\textbf{55.8}} & \gb{\textbf{57.8(+1.0)}} \\
    \midrule
    \multirow{5}{*}{DeiT-B/16} &  \gr{FPTrans~\cite{zhang2022feature}} & \gr{176.66M} &  \gr{44.4} &  \gr{48.9} &  \gr{50.6} & \gr{44.0} &  \gr{47.0} &  \gr{54.2} &  \gr{62.5} &  \gr{61.3} &  \gr{57.6} &  \gr{58.9} \\
    &  FPTrans* &176.66M&  43.0 &  49.6 &  48.0 & 43.2 &  46.0 &  54.5 &  63.6 &  59.8 &  56.9 &  58.7 \\
    & FPTrans + Adaptformer \gr{\shortcite{chen2022adaptformer}} & 0.49M & 43.7 & 50.0 & 48.4 & 44.1 & 46.6(+0.6) & 55.0 & \textbf{64.1} & 60.1 & 58.1 & 59.3(+0.6) \\
    & FPTrans + VPT\gr{\cite{jia2022visual}} & 0.25M & 43.8 & 50.1 & 48.5 & 43.7 & 46.5(+0.5) & 54.9 & 64.0 & 59.7 & 56.8 & 58.9(+0.2) \\
     & \gb{FPTrans + \textcolor{blue}{\textbf{Ours}}} & \gb{0.45M} & \gb{\textbf{45.3}} & \gb{\textbf{53.9}} & \gb{\textbf{49.5}} & \gb{\textbf{44.5}} & \gb{\textbf{48.3(+2.3)}} & \gb{\textbf{57.1}} & \gb{64.0} & \gb{\textbf{60.7}} & \gb{\textbf{58.2}} & \gb{\textbf{60.0(+1.3)}} \\
    \midrule
    \multirow{4}{*}{Swin-B} 
    
    & DCAMA\gr{~\cite{shi2022dense}} & 92.97M &  49.5 &  52.7 &  52.8 &  48.7 &  50.9 &  55.4 &  60.3 &  59.9 &  57.5 &  58.3 \\
   & DCAMA + F-Decoder & 5.07M & 49.6 & 52.4 & 52.7 & 51.0 & 51.4(+0.5) & 55.8 & 60.6 & 59.7 & 57.9 & 58.2(+0.2) \\
   & DCAMA + F-Head & 290 & 49.7 & 52.5 & 52.7 & 48.7 & 50.9(+0.0) & 55.4 & 60.4 & 59.8 & 57.8 & 58.4(+0.1) \\
     & \gb{DCAMA + \textcolor{blue}{\textbf{Ours}}} & \gb{0.23M} & \gb{\textbf{51.4}} & \gb{\textbf{53.1}} & \gb{\textbf{54.4}} & \gb{\textbf{51.9}} & \gb{\textbf{52.7(+1.8)}} & \gb{\textbf{57.8}} & \gb{\textbf{61.0}} & \gb{\textbf{60.9}} & \gb{\textbf{58.5}} & \gb{\textbf{59.6(+1.3)}} \\
    \bottomrule[1.2pt]
  \end{tabular}
    \caption{Comparison with state-of-the-art methods on PASCAL-5$^i$ and COCO-20$^i$. We report 1-shot and 5-shot results using the mIoU (\%). $*$ denotes our implemented result. The \gr{gray font} indicates the existing SOTA method.}
    \label{SOTA_result}
    \vspace{-10pt}
\end{table*}

\section{Experiments}
\label{Experiments}
\subsection{Implementation Details}

\subsubsection{Datasets and Metrics.} We evaluate our proposed Adaptive FSS on PASCAL-5$^i$ and COCO-20$^i$, which are two standard FSS benchmarks. PASCAL-5$^i$ consists of PASCAL VOC 2012 \cite{everingham2010pascal} and SBD \cite{hariharan2014simultaneous} datasets. It comprises 20 classes that are divided into training and testing splits of 15 and 5 classes. COCO-20$^i$ is created from COCO 2014 \cite{lin2014microsoft} dataset, which contains 80 classes. We divide them into four splits with each split of 60 classes for training and 20 classes for testing. For evaluation, we use the mean intersection over union (mIoU) to compare with previous methods.

\noindent \subsubsection{Training Details.} 
Our framework is implemented with PyTorch \cite{paszke2019pytorch} and trained on RTX 3090 GPUs with 1000 iterations. The cross-entropy loss is employed in our experiments and the batch size is set to 4 for PASCAL-5$^i$ and COCO-20$^i$. We adopt the SGD optimizer, with momentum of 0.9,  weight decay of 0.001, and learning rate of 0.01.
In each fold, two and six samples per novel class are randomly selected from the train set at 1-shot and 5-shot situations respectively for fine-tuning and avoiding image leakage. 
The results of each experiment are the average results obtained from five random sampling trainings.

\subsection{Comparison with State-of-the-Art Methods}
\subsubsection{Quantitative Results.} To comprehensively evaluate our approach, we conduct experiments on four few-shot segmentation networks (MSANet, HDMANet, FPTrans, and DCAMA), which adopt three popular backbones (ResNet, Vision Transformer, and Swin Transformer) as shown in Table \ref{SOTA_result}. It is worth emphasizing that we followed the test setting of DCAMA so that the performance of the other method is slightly different from that in the original paper. As expected, our method consistently improves the performance of existing FSS methods with different encoders on two benchmarks. For example, the result of DCAMA is respectively boosted by 3.1\% mIoU and 1.8\% mIoU in 1-shot and 5-shot settings on PASCAL-5$^i$. Besides, the different feature dimensions of multiple stages in various models result in an inconsistent number of parameters.

Moreover, we compare other fine-tuning strategies including finetune decoder (F-Decoder), finetune head (F-Head), and two Parameter-Efficient-Tuning methods (Adaptformer \cite{chen2022adaptformer} and VPT \cite{jia2022visual}), which are commonly used in Computer Vision. Since Adaptformer and VPT are well-designed for ViT, we conduct experiments on FPTrans. In particular, the classification head of FPTrans is based on similarity matching without learnable parameters, so the experiment of F-Decoder and F-Head is implemented on DCAMA. Our approach surpasses all fine-tuning strategies by a large margin. We further provide comparisons of different models in the \textbf{Appendix} \ref{More_SOTA_result}. 

\noindent \subsubsection{Results Analysis.} The large performance improvement can be explained by three factors. \textbf{(1)} \ul{Feature adaptation works better}. In the existing FSS method based on meta-learning, the classifier is not strongly associated with the specific category. The model relies on correlation matching between the support feature and the query feature at the pixel level to distinguish foreground from background. Therefore, the limited adaptability of F-Head results in a subtle improvement compared to Adaptformer, VPT, and F-Decoder. \textbf{(2)} \ul{The class information contained in the support feature and mask is beneficial to adaptation to novel classes}. In each fold, the multiple classes (i.e., 5 classes in PASCAL-5$^i$ and 20 classes in COCO-20$^i$) need to be adapted. An effective fine-tuning strategy should help the model extract class-specific features for different test classes by support set.
However, the Adaptformer and VPT only insert learnable parameters and modules to independently model task-specific features without utilizing the category information.
Our approach can greatly achieve this purpose through the well-designed PEM.
\textbf{(3)} \ul{Our method can memorize category-specific knowledge through the class prototype bank (CPB)}, obtaining high-quality representation for different categories. Ablation studies in Table \ref{components} confirm PEM and CPB as the main reasons for performance improvement.

\begin{table*}[t]
    \centering
    \footnotesize
    \begin{subtable}[t]{0.22\linewidth}
        \centering
        \renewcommand{\arraystretch}{1.3}
        \setlength{\tabcolsep}{0.7mm}{
        \begin{tabular}{c c c|c}
        % \toprule[0.75pt]
        LAE &  PEM & w/o CPB  & mIoU$\%$ \\
        \hline
           &  & -  & 68.8 \\
         \checkmark &  &   -  & 70.1 \\
        \checkmark & \checkmark &   \checkmark  & 71.4 \\
         \checkmark &  \checkmark &    & \ul{72.3} \\
        % \bottomrule[0.75pt]
	\end{tabular}}
        \caption{Major Components.}
        \label{components}
    \end{subtable}
    \begin{subtable}[t]{0.26\linewidth}
        \centering
        \renewcommand{\arraystretch}{1.3}
        \setlength{\tabcolsep}{0.7mm}{
        \begin{tabular}{c | c| c}
            % \toprule[0.75pt]
            Strategy  & Params  & mIoU$\%$ \\
            \hline
            % \hline
            1 $\longrightarrow$ 6  & 0.45M &  71.8 \\
            3 $\longrightarrow$ 8  & 0.45M & 71.6 \\
            7 $\longrightarrow$ 12  & 0.45M & \ul{72.3} \\
            1 $\longrightarrow$ 12 & 0.82M &  72.1 \\
            % \bottomrule[0.75pt]
        \end{tabular}}
        \caption{Insertion Strategy.}
        \label{Insert}
    \end{subtable}
    \begin{subtable}[t]{0.18\linewidth}
        \centering
        \renewcommand{\arraystretch}{1.3}
		\setlength{\tabcolsep}{2.2mm}{
		\begin{tabular}{c| c}
			% \toprule[0.75pt]
			  $\alpha$ & mIoU$\%$ \\
			\hline
                0.999 & 72.0 \\
                0.99  &  \ul{72.3} \\
                0.95  &  71.9 \\
                0.9 & 71.8 \\
			% \bottomrule[0.75pt]
		\end{tabular}}
            \caption{Momentum Ratio $\alpha$.}
        \label{Momentum}
    \end{subtable}
    \begin{subtable}[t]{0.16\linewidth}
        \centering
        \renewcommand{\arraystretch}{1.3}
		\setlength{\tabcolsep}{2.2mm}{
		\begin{tabular}{c| c}
			$\beta$   & mIoU$\%$ \\
			\hline
			1   & 66.7 \\
			0.5    & 69.2 \\
   		  0.1   &  \ul{72.3} \\
                0.01    & 71.0 \\
		\end{tabular}}
            \caption{Scaling Factor $\beta$.}  
		\label{Weight}
    \end{subtable}
    \begin{subtable}[t]{0.15\linewidth}
        \centering
        \renewcommand{\arraystretch}{1.3}
        \setlength{\tabcolsep}{2.2mm}{
        \begin{tabular}{c| c}
        $\gamma$  & mIoU$\%$ \\
        \hline
        4    & 71.8 \\
        8    & 71.9 \\
        16   &  \ul{72.3} \\
        32    &  71.5 \\
        \end{tabular}}
        \caption{Hidden Ratio $\gamma$.}
        \label{Hidden}
    \end{subtable}
    \caption{Ablation study with FPTrans on PASCAL-5$^i$. \textbf{The default settings of our method are marked in \ul{underline}}. }
    \vspace*{-0.4cm}
\label{ablation_study}
\end{table*}

\subsubsection{Qualitative Results.}
\begin{figure}[!t]
    \centering
    \includegraphics[width=0.45\textwidth]{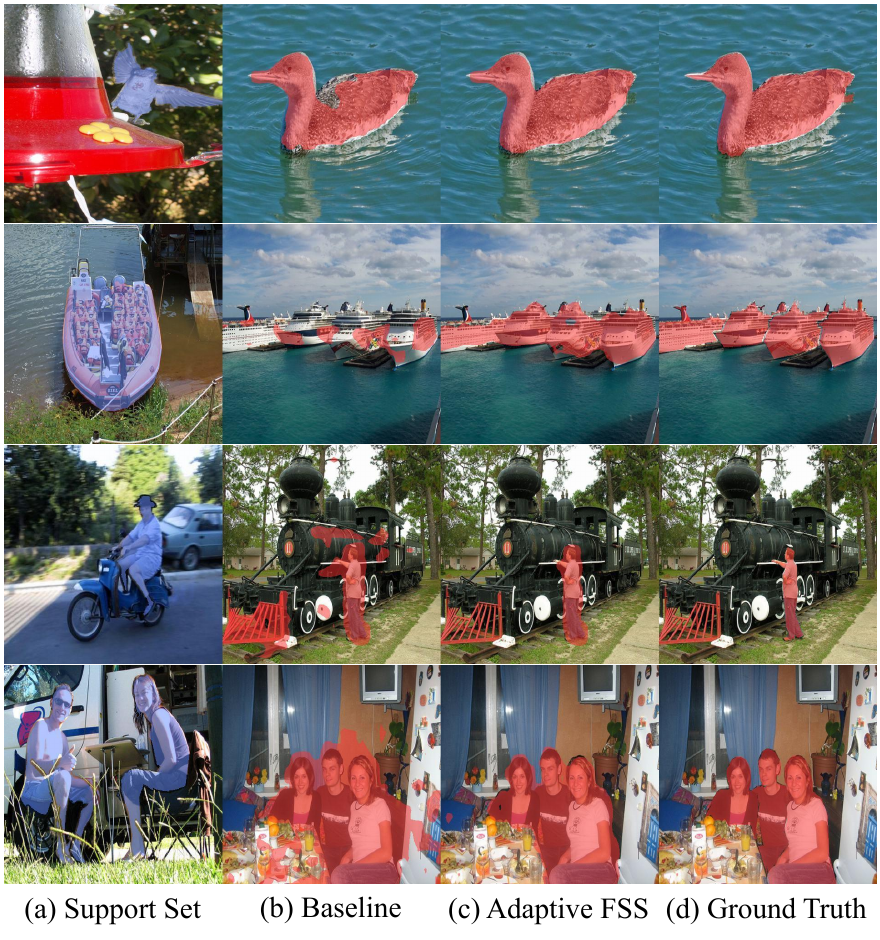}
    \caption{The visual comparison between baseline and our proposed Adaptive FSS on PASCAL-5$^i$ in 1-shot setting.}
    \label{visual_comparison}
    \vspace{-10pt}
\end{figure}

We further provide a visual comparison between the baseline and our Adaptive FSS on the PASCAL-5$^i$ as shown in Fig. \ref{visual_comparison}. The FPTrans without finetuning is chosen as the baseline. Our method achieves high-quality segmentation due to adapting the model to new categories effectively. More visualizations are available in the \textbf{Appendix} \ref{More_visual_result}.

\subsection{Ablation Study}
In this subsection, various ablation experiments are conducted on the PASCAL-5$^i$ with the FPTrans, which are summarized in Table \ref{ablation_study}. We verify the effectiveness of major components in PAM and explore the effect of the insert strategy, momentum ratio, scaling factor, and hidden ratio. 

\noindent \subsubsection{Major Components.} As shown in Table \ref{components}, we investigate the effectiveness of key components (i.e., the learnable adaptive module (LAM) and the prototype enhancement module (PEM)) in the designed PAM. Compared to 68.8\% mIoU without adaptation on novel classes, the performance is increased by 1.3\% mIoU when only using the LAM. 
Moreover, the result is further improved by 1.3\% mIoU, using the temporary prototype $P_t$ to guide subsequent enhancement operations without the class prototype bank (CPB). 
Finally, the full PAM promotes the performance to reach 72.3\% mIoU, which illustrates that the representation quality of the prototype contributes to precise segmentation.

\noindent \subsubsection{Insert Strategy.} As shown in Table \ref{Insert}, we take ViT as the backbone to explore the effect of PAM insertion position. 
Specifically, we explore four ways of inserting, including the front (i.e., `1 $\rightarrow$ 6' in Table \ref{Insert} denotes means inserting PAM in the layers of the model from 1 to 6), middle, back, and all layers of the model. The front and middle approaches achieved similar results. Inserting PAM at the back of the model outperforms others. The reason is that the high-level semantic information, which is conducive to prototype extraction, is located in the deep layer of the network. Higher-quality object prototypes enable more precise association and feature augmentation. 
Moreover, this characteristic is more prominent for DCAMA, MSANet, and HDMNet based on the hierarchical Swin Transformer and ResNet. 
Therefore, for the former, we evenly insert the PAM in the last two stages. Due to the different structures between the ResNet-50 and Swin-B, the PAM is only plugged into the last stage at the ResNet-50. Relevant experimental results and more detailed explanations are available in the \textbf{Appendix} \ref{more_ab}.

\noindent \subsubsection{Momentum Ratio.} As described in Section \ref{Prototype_Enhancement_Module}, the momentum ratio $\alpha$ is introduced to control the rate of updating prototypes in the class prototype bank. We conduct experiments with the different $\alpha$ as shown in Table \ref{Momentum}. We found that $\alpha=0.99$ performs the best and set it as the default.

\noindent \subsubsection{Scaling Factor.} The scaling factor $\beta$ is introduced to balance the original features and task-specific features. We evaluate the performance when selecting different $\beta$ as shown in Table \ref{Weight}. Obviously, $\beta$ = 0.1 achieves the best results. Increasing or decreasing $\beta$ will bring a performance drop. Therefore, we chose 0.1 as the default setting.

\noindent \subsubsection{Hidden Ratio.} As mentioned in Section \ref{Learnable_Adaptive_Module}, the hidden ratio $\gamma$ can control the number of parameters introduced by PAM. A lower ratio means that more parameters are introduced and the task-specific ability is stronger. We study the impact of this hyper-parameter on the performance of the model as shown in Table \ref{Hidden}. It is observed that using 16 as a hidden ratio achieves the highest accuracy on PASCAL-5$^i$ for our method. Increasing $\gamma$ from 16 to 32 results in a performance drop from 72.3\% to 71.5\% mIoU. 

\section{Discussion}
\label{Discussion}
In this section, we further discuss the effectiveness of Adaptive FSS from three aspects as follows. 
\subsubsection{Strong adaptation ability of Adaptive FSS.}
 In Table \ref{transfer_result}, we further explore the adaptability of proposed Adaptive FSS for cross-domain segmentation, following the experimental setup in previous works \cite{zhang2022feature, min2021hypercorrelation}. 
 We fine-tune and test FPTrans on new classes in PASCAL-5$^i$, utilizing weights trained on base classes in COCO-20$^i$. 
 Due to the difference in the hardware experiment environment, our realized results (68.2\% mIoU) are different from 69.7\% mIoU in the original paper.
 Our method obtains significant improvement from 68.2\% to 72.5\% mIoU and exhibits strong superiority in the presence of domain gaps.

\begin{table}[htbp]
\centering
\footnotesize
    \renewcommand{\arraystretch}{1.3}
    \setlength{\tabcolsep}{1.2mm}{
    \begin{tabular}{c| c c c}
        \toprule[1.0pt]
        COCO-20$^i$ to PASCAL-5$^i$   & FPTrans$\dagger$ & FPTrans  & \gb{Ours} \\
        \hline
        Mean IoU$\%$ & 68.2 & 69.7 &  \gb{\textbf{72.5 (+4.3)}} \\
        \bottomrule[1.0pt]
    \end{tabular}}
    \caption{Evaluation under the domain shift from COCO-20$^i$ to PASCAL-5$^i$. $\dagger$ denotes our realized result.}
    \label{transfer_result}
    \vspace{-10pt}
\end{table}

\subsubsection{Only One Labeled Sample per Class Is Available for Fine-Tuning.}
% It is an extreme situation. 
To cope with this problem, we explore a compromised training method using only one image simultaneously as the query and support image. Specifically, we utilize the original sample as query and the data-augmented sample as support for fine-tuning. Among them, data enhancement methods such as random rotation, random cropping, horizontal flipping, and vertical flipping are adopted. As shown in Table \ref{one-sample}, we use FPTrans as the baseline and experiment on PASCAL-5$^i$ in 1-shot setting. Our method achieves an improvement of 1.6\% mIoU, which demonstrates the superiority and robustness of Adaptive FSS.

\begin{table}[htbp]
    \centering
    \footnotesize
    \renewcommand{\arraystretch}{1.3}
    \setlength{\tabcolsep}{1.5mm}{
    \begin{tabular}{c |c c c c |c}
        \toprule[1.0pt]
        Method & Fold-0 & Fold-1 & Fold-2 & Fold-3 & Mean IoU$\%$ \\
        \hline
        FPTrans  & 73.4 & 70.3 & 67.8 & 63.7   & 68.8 \\
        \rowcolor{gray!18} Ours  & \textbf{73.8} & \textbf{73.2} & \textbf{69.4} & \textbf{65.0}    & \textbf{70.4 (+1.6)} \\
        \bottomrule[1.0pt]
    \end{tabular}}
    \caption{Evaluation under only one labeled sample.}
    \label{one-sample}
    \vspace{-10pt}
\end{table}

\noindent \subsubsection{The Reason for Performance Improvement.} It is the effectiveness of Adaptive FSS rather than extra labeled samples. To prove it, we prevent the influence of additional labeled images by reusing the fine-tuning set as the support set during inference.
Specifically, two samples per class act as query and support alternately in fine-tuning, constituting the fine-tuning set. In the inference phase, they are both as support in each inference, which is a 2-shot test setting.
Meanwhile, DCAMA employs a 1-shot training and few-shot testing manner, it can flexibly adapt to different numbers of support images. Therefore, we adopt DCAMA as the baseline. As evident from Table \ref{two_shot}, our method achieves a result of 75.0 \% mIoU with an improvement of 2.4\% mIoU.

\begin{table}[htbp]
    \centering
    \footnotesize
    \renewcommand{\arraystretch}{1.3}
    \setlength{\tabcolsep}{1.5mm}{
    \begin{tabular}{c |c c c c |c}
        \toprule[1.0pt]
        Method & Fold-0 & Fold-1 & Fold-2 & Fold-3 & Mean IoU$\%$ \\
        \hline
        DCAMA  & 72.8 & 75.2 & 69.3 & 73.3   & 72.6 \\
        \rowcolor{gray!18} Ours  & \textbf{74.6} & \textbf{75.9} & \textbf{74.9} & \textbf{74.4}    & \textbf{75.0 (+2.4)} \\
        \bottomrule[1.0pt]
    \end{tabular}}
    \caption{Evaluation on the influence of extra labeled sample.}
    \label{two_shot}
    \vspace{-10pt}
\end{table}

\section{Conclusion}
In this paper, we propose a novel FSS framework based on the adapter mechanism that can greatly improve the performance of the current FSS model by adapting it to novel classes. The well-designed PAM could accurately guide the enhancement of features with a high correlation to new objects. The experiments verify that Adaptive FSS can achieve a considerable improvement on diverse powerful FSS networks based on a variety of backbones. We also explore the performance of our method in the presence of domain gaps and only one training sample. Furthermore, we demonstrate that the superiority of Adaptive FSS is independent of additional samples. We hope that our approach can serve as a strong baseline for the novel class adaptation of FSS in future research.

\section*{Acknowledgements}
This work was supported by National Key Research and Development Program of China (2022YFB3304000), in part by the Natural Science Foundation of China under Grant 42201386, in part by the Fundamental Research Funds for the Central Universities and the Youth Teacher International Exchange and Growth Program of USTB (QNXM20220033), and Interdisciplinary Research Project for Young Teachers of USTB (Fundamental Research Funds for the Central Universities: FRF-IDRY-22-018), and Scientific and Technological Innovation Foundation of Shunde Innovation School, USTB (BK20BE014).

\bibliography{aaai24}

\clearpage
% \section{Conclusion}
\appendix
\twocolumn[{%
\renewcommand\twocolumn[1][]{#1}%
\section*{Appendix}

\vspace{20.pt}
In this Appendix, we provide the following items:
\begin{enumerate}
    \item (Sec. \ref{More_SOTA_result}) Quantitative results based on MSANet~\cite{iqbal2022msanet} and HDMNet~\cite{peng2023hierarchical} on the PASCAL-5$^i$ \cite{everingham2010pascal} and COCO-20$^i$ \cite{lin2014microsoft}.).
    \item (Sec. \ref{More_visual_result}) Visual comparison of segmentation results for qualitative analysis.
    \item (Sec. \ref{more_ab}) More ablation studies on insert strategy at various backbone, including ResNet-50 and Swin Transformer base.
    \item (Sec. \ref{more_transfer_result}) More experimental results about domain shift based on DCAMA\cite{shi2022dense}.
\end{enumerate}

\begin{center}
    \centering
  \footnotesize
  \setlength\tabcolsep{1.6pt}%
  \begin{tabular}{c|l|c|cccc|c|cccc|c}
    \toprule[1.0pt]
    \multicolumn{13}{c}{\textbf{(a) PASCAL-5$^i$}}  \\
    \multirow{2}{*}{Backbone} & \multirow{2}{*}{Method} & \multirow{2}{*}{Params} & \multicolumn{5}{c|}{1-shot} & \multicolumn{5}{c}{5-shot} \\
    \cmidrule{4-13}
      &  & & fold-0 & fold-1 & fold-2 & fold-3 & mIoU\% & fold-0 & fold-1 & fold-2 & fold-3 & mIoU\% \\
    \midrule

        \multirow{7}{*}{ResNet 50} 

     &  \gr{HDMNet~\cite{peng2023hierarchical}} & \gr{51.40M} &
     \gr{71.0} &  \gr{75.4} & \gr{68.9} &  \gr{62.1}&
     \gr{69.4} & \gr{71.3} &  \gr{76.2} &  \gr{71.3} &  \gr{68.5} &  \gr{71.8} \\

     &  \gr{MSANet~\shortcite{iqbal2022msanet}} & \gr{52.37M} & \gr{69.3} &  \gr{74.6} & \gr{67.8} &  \gr{62.4} & \gr{68.5} & \gr{72.7} &  \gr{76.3} &  \gr{73.5} &  \gr{67.9} &  \gr{72.6} \\
     &  MSANet* & 52.37M & 69.9 &  74.9 & 64.8 &  61.3 & 67.7 & 74.0 &  76.4 &  69.8 &  68.0 &  72.1 \\
      &  MSANet + F-Decoder & 5.13M & 70.8 & 74.7 & 64.7 & 62 & 68.1(+0.4) & 73.6 & 76.2 & 72.2 & 67.4 & 72.4(+0.3) \\
      &  MSANet + F-Head & 518 & 70.8 & 74.7 & 64.7 & 61.5 & 67.9(+0.2) & 73.6 & 76.3 & 71.1 & 67.2 &  72.1(+0.0) \\
      &  MSANet + Conv-Adapter (Conv. Par.) & 1.30M & 70.7 & 74.9 & 64.9 & 61.8 & 68.1(+0.4) & 74.1 & 76.5 & 71.8 & 68.3 & 72.7(+0.6) \\

     & \gb{MSANet + \textbf{Ours}} & \gb{0.53M} & \gb{\textbf{71.1}} & \gb{\textbf{75.5}} & \gb{\textbf{67.0}} & \gb{\textbf{64.5}} & \gb{\textbf{69.5(+1.8)}} & \gb{\textbf{74.7}} & \gb{\textbf{78.0}} & \gb{\textbf{75.3}} & \gb{\textbf{70.8}} & \gb{\textbf{74.7(+2.6)}} \\

  \end{tabular}
  \setlength\tabcolsep{1.6pt}%
      \begin{tabular}{c|l|c|cccc|c|cccc|c}
    \toprule[1.0pt]
    \multicolumn{13}{c}{\textbf{(b) COCO-20$^i$}}  \\
    \multirow{2}{*}{Backbone} & \multirow{2}{*}{Method} & \multirow{2}{*}{Params} & \multicolumn{5}{c|}{1-shot} & \multicolumn{5}{c}{5-shot} \\
    \cmidrule{4-13}
      & & & fold-0 & fold-1 & fold-2 & fold-3 &  mIoU\% & fold-0 & fold-1 & fold-2 & fold-3 &  mIoU\% \\
    \midrule
    \multirow{6}{*}{ResNet 50}

    &  \gr{HDMNet~\cite{peng2023hierarchical}} & \gr{51.40M} & 
     \gr{43.8} &  \gr{55.3} & \gr{51.6} &  \gr{49.4}&
     \gr{50.0} & \gr{50.6} &  \gr{61.6} &  \gr{55.7} &  \gr{56.0} &  \gr{56.0} \\
     &  HDMNet* & 51.40M & 44.0  & 55.1  & 50.1 &  48.7 &  49.5 & 52.5  & 64.5 &  55.2 &  55.1 &  56.8 \\
     &  HDMNet + F-Decoder & 3.55M & 44.1 &  55.1 & 48.9 & 48.8 & 49.2(-0.3) & 53.2 & 65.1 & 55.5 & 55.1 & 57.2(+0.4) \\
    &  HDMNet + F-Head & 132 & 44.8 & 55.1 & 50.2 &  48.7 & 49.7(+0.2) & \textbf{53.6} & 64.8 &  55.0 & 55.1 & 57.1(+0.3) \\
    &  HDMNet + Conv-Adapter (Conv. Par.) & 1.30M & 44.5 & 55.8 & 50.8 & 48.8 & 50.0(+0.5) & 53.2 &  65.2 & 55.5 & 55.4 & 57.3(+0.5) \\
    & \gb{HDMNet + \textbf{Ours}}  & \gb{0.53M} & \gb{\textbf{44.9}} & \gb{\textbf{56.7}} & \gb{\textbf{51.4}} & \gb{\textbf{49.6}} & \gb{\textbf{50.7(+1.2)}} & \gb{53.0} & \gb{\textbf{66.4}} & \gb{\textbf{56.1}} & \gb{\textbf{55.8}} & \gb{\textbf{57.8(+1.0)}} \\
    \bottomrule[1.2pt]
  \end{tabular}
    \captionof{table}{Comparison with state-of-the-art methods on PASCAL-5$^i$ and COCO-20$^i$. We report 1-shot and 5-shot results using the mean IoU (\%). $*$ denotes our implemented result. The \gr{gray font} indicates the existing SOTA method.}
    \label{SOTA_result_app}
\end{center}% 
}]

\section{Quantitative Results}
\label{More_SOTA_result}
In Table \ref{SOTA_result_app}, we evaluate fine-tuning strategies (i.e.,  fine-tuning decoder (F-Decoder) and fine-tuning head (F-Head)) on MSANet and HDMNet. At the same time, Conv-Adapter \cite{chen2022conv}, a parameter-efficient-tuning method specially designed for ResNet \cite{he2016deep}, is employed for a comprehensive comparison. To address the challenge of limited samples in this task, we select Conv. Par., which has the fewest parameters among the four Conv-Adapter approaches, to conduct experiments. On the two FSS methods based on ResNet-50, our method significantly improves the performance compared to other fine-tuning strategies as shown in Table \ref{SOTA_result_app}. The results demonstrate the generality and superiority of our approach.

\section{Qualitative Results}
\label{More_visual_result}
\begin{figure*}[t]
    \centering
    \includegraphics[width=0.95\textwidth]{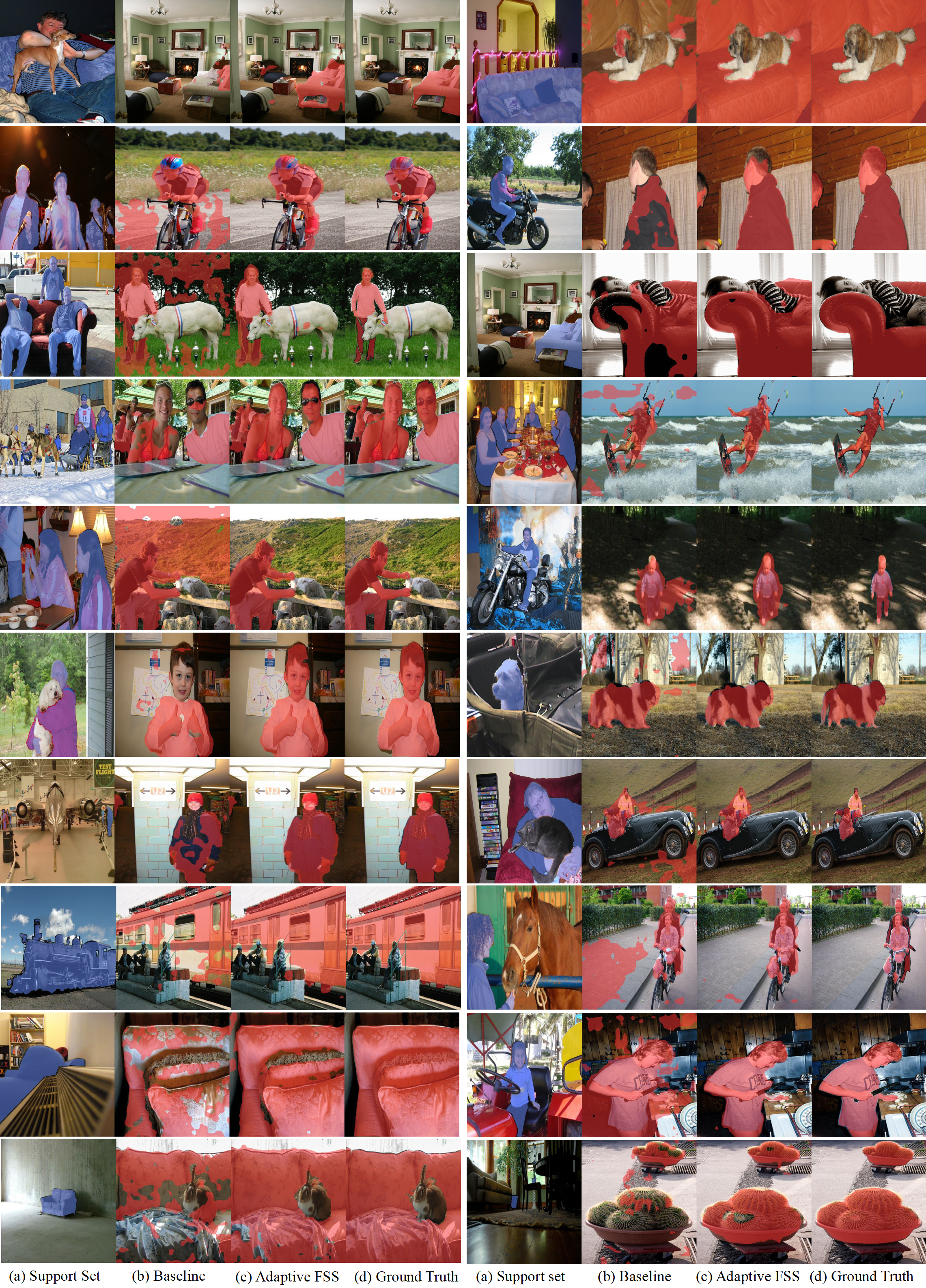}
    \caption{The visual comparison between baseline and our proposed Adaptive FSS on PASCAL-5$^i$ in 1-shot setting.}
    \label{visual_comparison_app}
\end{figure*}

The segmentation results of Adaptive FSS on the PASCAL-5$^i$ dataset are shown in Fig. \ref{visual_comparison_app}. We use the FPTrans \cite{zhang2022feature} without finetuning as the baseline for comparison. 
It is obvious that Adaptive FSS achieves accurate segmentation of test-class targets compared to baseline.
This illustrates that adapting FPTrans to novel classes by Adaptive FSS enables the capturing of sharper object boundaries and reduces the presence of erroneous segmenting.

\section{More Ablation Studies on Insert Strategy}
\label{more_ab}
To explore the impact of different numbers and insertion positions of PAM, we conducted abundant experiments on the DCAMA and MSANet, which are based on the hierarchical backbone Swin-B and ResNet-50. The number of layers for each stage in ResNet-50 and Swin-B is (3, 4, 6, 3) and (2, 2, 18, 2).
\begin{table}[!h]
  \centering
  \footnotesize
  \renewcommand{\arraystretch}{0.5}
  \setlength\tabcolsep{0.5pt}%
  \begin{tabular}{c|cccc|c|c}
    \toprule
    \multirow{2}{*}{\makecell[c]{Backbone \\  \& Method}} & \multicolumn{4}{c|}{Insert Strategy}  & \multirow{2}{*}{Params}& \multirow{2}{*}{mIoU\%} \\
    \cmidrule{2-5}
     & Stage 1 & Stage 2 & Stage 3 & Stage 4  &  & \\
    \midrule

        \multirow{8}{*}{\makecell[c]{ResNet-50 \\ \& MSANet}} 

     &  \gr{-} & \gr{-} & \gr{-} & \gr{-} & \gr{52.37M}  & \gr{67.7} \\
     &  1$\rightarrow$3 & 1$\rightarrow$4 & 1$\rightarrow$6 & 1$\rightarrow$3  & 2.53M &   67.9 (+0.2) \\
     &  1$\rightarrow$3 & 1$\rightarrow$4 & - & - & 0.15M &    67.6 (-0.1)\\
     &  - & - & 1$\rightarrow$6 & 1$\rightarrow$3  & 2.37M &   68.5 (+0.8)\\
     &  - & - & 1$\rightarrow$6 & - & 0.79M &    68.4 (+0.7)\\
     &  - & - & - & 1$\rightarrow$3  & 1.57M &   \textbf{69.8 (+2.1)}\\
     &  \gb{-} & \gb{-} & \gb{-} & \gb{\{1\}} & \gb{0.53M} &  \gb{69.5 (+1.8)}\\
     &  - & - & - & \{2\} & 0.53M &  69.7 (+2.0)\\

    \midrule
    \multirow{8}{*}{\makecell[c]{Swin-B \\  \& DCAMA}} 
    & \gr{-} & \gr{-} & \gr{-} & \gr{-} & \gr{92.97M} &  \gr{69.3} \\
    &  1$\rightarrow$2 & 1$\rightarrow$2 & 1$\rightarrow$18 & 1$\rightarrow$2& 0.88M & 70.4 (+1.1)\\
    & 1$\rightarrow$2 & 1$\rightarrow$2 & - &  - & 0.02M &  69.7 (+0.4)\\
    &   - &  - & 1$\rightarrow$18 & 1$\rightarrow$2   & 0.86M &  72.1 (+2.8)\\
    &    - &  - & \{1,3,...,17\} & \{1\} & 0.43M &   \textbf{72.6 (+3.3)}\\
    &  \gb{-} & \gb{-} & \gb{\{1,7,13\}} & \gb{\{1\}}  & \gb{0.23M}  &  \gb{72.4 (+3.1)}\\
    &   - &  - & \{6,12,18\} & \{2\}   & 0.23M  &  72.3 (+3.0)\\
    & - & - & \{1\} & \{1\} & 0.16M  &   69.9 (+0.6)\\

    \bottomrule
  \end{tabular}
      \vspace{-8pt}
    \small
    \caption{Ablation study on the insert strategy of PAM. \{1\} and 1$\rightarrow$6 denote inserting the PAM in layer 1 and layers 1 to 6 respectively.}
    \label{ablation_insert}
    \vspace{-10pt}
\end{table}

The experimental results are exhibited in Table \ref{ablation_insert} and can be analyzed from three aspects. 
\textbf{1)} The acquisition of high-quality class prototypes in PAM requires input features with rich semantic information. Intuitively, adopting lower-resolution feature maps in deep layers of the network are reasonable choice. On the contrary, the performance improvement is slight when only inserting PAM in shallow stages, such as stages 1-3 for ResNet-50 and stages 1-2 for Swin-B.
\textbf{2)} Although inserting PAM in all layers of the selected stage can improve performance impressively, an appropriate sparse insertion strategy is superior by achieving the trade-off between parameter efficiency and accuracy.
Specifically, using the insertion strategy of (-, -, -, \{1\}) lead to competitive results 69.5\% mIoU and fewer learnable parameters 0.53M compared to 69.8\% mIoU and 1.57M using the method of (-, -, -, 1$\rightarrow$3).
\textbf{3)} When the number and stage of PAM insertions are fixed, changing the insertion position has less impact on performance. For instance, the results of 72.3\% mIoU and 72.4\% mIoU obtained by schemes (-, -, \{1, 7, 13\}, \{1\}) and (-, -, \{6, 12, 18\}, \{2\}) are close in Table \ref{ablation_insert}. 

\textcolor{magenta}{\textbf{In general, reasonable and sparse insertion in the deeper stage is recommended when using the Adaptive FSS}}.

\section{More Results About Domain Shift}
\label{more_transfer_result}
We further explored
the performance of DCAMA with Adaptive FSS on the domain shift task. As expected, our method achieves outstanding improvements (i.e., 7.0 mIoU\%) as shown in Table \ref{transfer_result_dcama}.
\begin{table}[htbp]
\centering
\footnotesize
    \renewcommand{\arraystretch}{1.3}
    \setlength{\tabcolsep}{1.2mm}{
    \begin{tabular}{c| c c}
        \toprule
        COCO-20$^i$ to PASCAL-5$^i$   & DCAMA  & \gb{DCAMA + Ours} \\
        \hline
        Mean IoU$\%$ & 66.0 &  \gb{\textbf{73.0 (+7.0)}} \\
        \bottomrule
    \end{tabular}}
    \caption{Evaluation under the domain shift from COCO-20$^i$ to PASCAL-5$^i$. $\dagger$ denotes our realized result.}
    \label{transfer_result_dcama}
    \vspace{-10pt}
\end{table}

\end{document}